\documentclass[letterpaper, 10 pt, conference]{ieeeconf}  

\IEEEoverridecommandlockouts                              

\DeclareUnicodeCharacter{0301}{*************************************}

\usepackage{amsmath} 
\usepackage{xcolor}
\usepackage{graphicx}
\usepackage{tabularx}
\usepackage{booktabs} 
\usepackage{multirow}
\usepackage{graphicx} 
\usepackage{caption,subcaption}
\usepackage{algorithm}
\usepackage{bm}
\usepackage{algpseudocode}

\title{\LARGE \bf
Asymptotically-Bounded 3D Frontier Exploration enhanced with Bayesian Information Gain
}

\author{John Lewis$^{1}$, Meysam Basiri$^{1}$, and Pedro U. Lima$^{1}$
\thanks{* This work was supported by doctoral grant from Fundação para a Ciência e a Tecnologia (FCT) UI/BD/153758/2022, Aero.Next project (PRR - C645727867-00000066) and ISR/LARSyS Strategic Funding through the FCT project
DOI: 10.54499/UIDB/50009/2020, DOI: 10.54499/UIDP/50009/2020, DOI: 10.54499/LA/P/0083/2020}
\thanks{$^{1}$J. Lewis, M. Basiri and P. Lima are with the Institute for Systems and Robotics,
Instituto Superior T\'ecnico, Universidade de Lisboa, Lisbon 1049-001, Portugal
(e-mail: \{\textit{john.lewis; meysam.basiri; pedro.lima}\}@\textit{tecnico.ulisboa.pt}).}%
}

\usepackage[
backend=biber,
style=numeric-comp,
sorting=none
]{biblatex}
\bibliography{IEEEexample}

\usepackage{makecell}
\begin{document}

\maketitle
\thispagestyle{empty}
\pagestyle{empty}

\begin{abstract}
Robotic exploration in large-scale environments is computationally demanding due to the high overhead of processing extensive frontiers. This article presents an OctoMap-based frontier exploration algorithm with predictable, asymptotically bounded performance. Unlike conventional methods whose complexity scales with environment size, our approach maintains a complexity of $\mathcal{O}(|\mathcal{F}|)$, where $|\mathcal{F}|$ is the number of frontiers. This is achieved through strategic forward and inverse sensor modeling, which enables approximate yet efficient frontier detection and maintenance. To further enhance performance, we integrate a Bayesian regressor to estimate information gain, circumventing the need to explicitly count unknown voxels when prioritizing viewpoints. Simulations show the proposed method is more computationally efficient than the existing OctoMap-based methods and achieves computational efficiency comparable to baselines that are independent of OctoMap. Specifically, the Bayesian-enhanced framework achieves up to a $54\%$ improvement in total exploration time compared to standard deterministic frontier-based baselines across varying spatial scales, while guaranteeing task completion. Real-world experiments confirm the computational bounds as well as the effectiveness of the proposed enhancement.

\end{abstract}
\section{INTRODUCTION} \label{sec:introduction}

\label{sec:literature_review}
Autonomous exploration is fundamental to operating in unknown areas where uncertainty, redundancy, and safety concerns discourage manual effort. By combining path planning with simultaneous localization and mapping (SLAM), autonomous systems enable the quick coverage of large-scale environments for search and rescue operations, 3D reconstruction, and structural inspections \cite{10582913, zhang2024soar, zhang2024falcon, feng2024fc}. However, while autonomous exploration is highly effective, state-of-the-art algorithms often overlook the significant computational and memory overheads amplified in 3D settings. As the exploration volume increases, processing these expansive spaces can severely degrade performance, making it crucial to decouple computational efficiency from the total map volume.

To navigate these environments, exploration strategies are typically classified based on viewpoint selection. A frontier-based exploration strategy \cite{batinovic2021multi,EPIC} aims to process the boundary between the known and unknown (frontiers), while sampling-based strategies \cite{batinovic2022shadowcasting} utilize a randomized approach to generate candidate viewpoints and trajectories. Hybrid approaches \cite{ribeiro2024efficient,zhang2024falcon} fuse these methods to determine the next best viewpoint. Furthermore, the planner can choose to visit viewpoints conservatively one after another \cite{batinovic2021multi} or batch a subset of viewpoints as a trajectory \cite{EPIC}. Recent lightweight frameworks \cite{EPIC} expand on the batched approach by clustering detected surface frontiers and formulating an Asymmetric Traveling Salesman Problem (ATSP) to calculate a global guidance path that visits candidate viewpoints in a highly energy-efficient sequence.

Once a viewpoint is determined, frameworks employ motion planning algorithms like A* \cite{zhang2024falcon} or RRT \cite{10582913} on a 3D representation of the environment—such as TSDF \cite{oleynikova2016signed}, OctoMap \cite{wurm2010octomap}, UFO \cite{duberg2020ufomap}, or Observation maps \cite{EPIC}—to determine safe traversal paths. OctoMap \cite{wurm2010octomap}, for instance, leverages the efficient 3D space partitioning of Octrees and an inverse sensor model to determine voxel occupancy probabilistically. With a data access complexity of $\mathcal{O}(log(m))$ for $m$ nodes, OctoMap is a strong choice for resource-limited operations. Consequently, researchers have extensively studied its use in 3D exploration to optimize exploration time \cite{zhou2021fuel} and computational requirements \cite{batinovic2021multi}.

Despite efficient data structures, the key challenge in large-scale autonomous exploration remains the asymptotic time complexity of detecting and processing frontiers. In a trivial OctoMap approach, iterating across all leaf nodes to find free voxels with unknown neighbors incurs a complexity of $\mathcal{O}(m)$, which scales poorly as the map grows. To tackle this, modern methods extract frontiers incrementally \cite{batinovic2021multi}, tracking only changed cells ($\underbar{c}$ and splitting frontiers into local and global sets \cite{mannucci2017autonomous}. While this bounds the overall frontier computation to $\mathcal{O}(\underbar{c}+|\mathcal{F}|)$, where $|\mathcal{F}|$ is the cardinality of the frontier set, these methods remain vulnerable to severe computational spikes when observing massive new spaces. Conversely, frameworks operating directly on point clouds achieve real-time efficiency through localized updates—such as bounding overhead strictly to newly acquired data \cite{EPIC}—but rely on aggressive local heuristics that risk losing complete volumetric guarantees.

Beyond the computational overhead of detecting frontiers, evaluating their utility poses an additional challenge. Standard exploration frameworks (e.g., information gain cubes \cite{batinovic2021multi} or coverage scores \cite{EPIC}) typically rely on deterministic calculations, such as counting unknown/uncertain voxels/frontiers within a predefined threshold centered at a candidate viewpoint or robot pose. While computationally straightforward, this geometric approximation is often constrained and inaccurate. It can lead to suboptimal viewpoint selection, particularly in complex, unstructured environments where the spatial distribution of unknown voxels is highly non-uniform.

To bridge these gaps, we present an asymptotically-bounded exploration algorithm that addresses both frontier detection scaling and utility evaluation. First, we integrate frontier detection directly into the forward sensor model, shifting the computational complexity to strictly $\mathcal{O}(|\mathcal{F}|)$. This completely eliminates the dependency on updated cells, ensuring predictable computation times even in expansive environments without sacrificing volumetric guarantees. Second, rather than relying on rigid deterministic counts, our method integrates a lightweight Bayesian regressor. By modeling information gain as a probabilistic regression problem, the framework learns the exploratory utility of local voxel distributions, accelerating exploration and improving viewpoint selection without sacrificing processing speed.

\subsection{Contributions} \label{sec:contributions} 
This paper presents an OctoMap-based frontier exploration algorithm, designed to overcome computational bottlenecks in large-scale 3D environments. Our key contributions are:

\begin{itemize}
    \item \textbf{Asymptotically-Bounded Frontier Detection:} A novel 3D frontier detection and maintenance approach with a proven $\mathcal{O}(|\mathcal{F}|)$ complexity, ensuring predictable, bounded computation times in expansive environments.
    \item \textbf{Bayesian Information Gain Estimator:} A computationally lightweight Bayesian regressor that estimates future information gain, accelerating overall exploration without sacrificing processing speed.
    \item \textbf{Extensive Evaluation:} Software-In-The-Loop (SITL) simulations demonstrate superior computational efficiency and faster exploration than state-of-the-art baselines across diverse topologies. Successful real-world aerial deployments further confirm the system's practical robustness.
\end{itemize}


The article is structured as follows: Section \ref{sec:methodology} provides a detailed overview of our proposed OctoMap-based frontier exploration method, highlighting the key algorithmic innovations that ensure  $\mathcal{O}(|\mathcal{F}|)$ computational efficiency and enhancing viewpoint selection with Bayesian regressor based information gain estimation. The experimental setup and a comprehensive analysis of the results are presented in Section \ref{sec:results}. Finally, Section \ref{sec:conclusion} summarizes our findings and outlines potential avenues for future improvements to the proposed method.

\section{METHODOLOGY}
\label{sec:methodology}

\begin{figure}[ht]
  \centering
  \begin{minipage}[h]{0.5\textwidth}
      \centering
      \includegraphics[width=\textwidth]{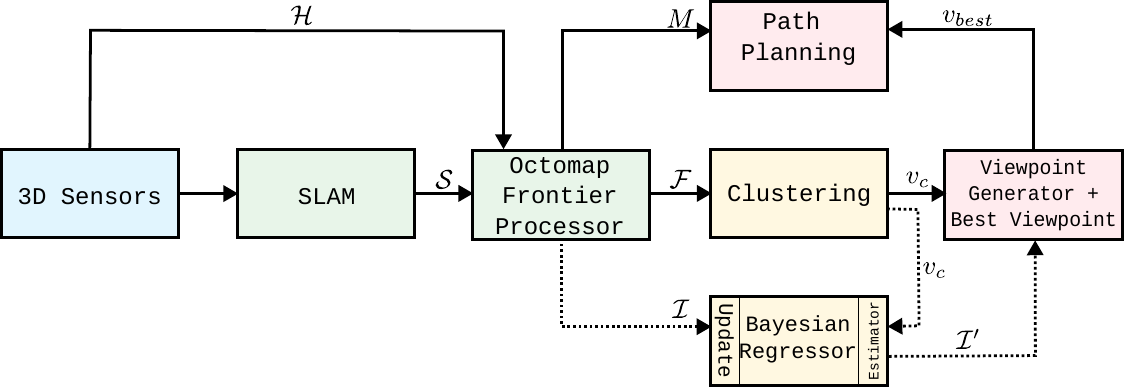}
     \caption{Proposed exploration framework utilizing frontier approximation to ensure asymptotic viewpoint generation. The dotted lines represent the proposed enhancement assisted by Bayesian Regressor for  information gain estimation.}
      \label{fig:Methodology}
	\end{minipage}
\end{figure}

Consider a robot, $\mathcal{R}$, equipped with one or more $3D$ perception sensors and tasked with exploring an unknown 3D volume, $V \subset \mathbf{R}^3$.  The volume $V$ can be divided into voxels, with each voxel belonging to one of the $3$ mutually exclusive subsets of $V_{free}$, $V_{o}$, and $V_{unknown}$ representing the free, occupied, and unknown volumes. The exploration task is considered to be a success when $V_{unknown}$ is minimized to a threshold. The exploration framework with the enhancement module is illustrated in Fig.\ref{fig:Methodology}, and the key components are detailed in the following sections.

\subsection{SLAM}

Inspired by the multi-resolution frontier-based 3D exploration planner proposed in  \cite{batinovic2021multi}, we generate the OctoMap from the SLAM sub-map rather than directly from raw sensor data. Sampling points from the SLAM map introduces an additional layer of modularity, enabling the fusion of multiple perception sensors to construct the OctoMap. This approach also couples the OctoMap update rate to the SLAM update rate—rather than the higher-frequency sensor update rate, thereby reducing the overall computational load. Moreover, compared to raw sensor point clouds, the sensor-fused SLAM output is more robust and less susceptible to measurement errors.

\subsection{OctoMap and Frontier Update}\label{subsection:octomapperfrontierprocessor}

OctoMap follows the Bayesian update rule to calculate the occupancy probability $P(n|z_{1:t})$, for the leaf node $n$ 
\begin{equation} \label{eq:OccupancyProbability} 
\left[ 1 + \frac{1-P(n|z_t)}{P(n|z_t)}  \frac{1-P(n|z_{1:t-1})}{P(n|z_{1:t-1})} \frac{P(n)}{1-P(n)}\right]^{-1}
\end{equation}
where $z_{1:t}$ are the sensor measurements. Using the log-odds notation and a uniform prior probability assumption of $P(n) = 0.5$, the authors of  \cite{hornung13auro} further reduce Eq. \ref{eq:OccupancyProbability} to 
\begin{equation} \label{eq:OccupancyProbabilityMinimal}
L(n|z_{1:t}) = L(n|z_{1:t-1}) + L(n|z_{t})
\end{equation}
with 
$L(n) = log\left[\frac{P(n)}{1-P(n)}\right]$. With no loss in generality, the occupancy probability $P(n|z_{1:t})$ of a voxel, $c$, in OctoMap $M$ is represented as $p_{\mathrm{o}}(c)$.  

\subsubsection{OctoMap Update}

In this work, we utilize a forward-sensor model to detect the frontiers in conjunction with the OctoMap's inverse sensor model. The inverse sensor model updates voxel probabilities given measurements, while the forward sensor model predicts expected measurements given the current map and sensor pose. We combine both: inverse for updating, forward for detecting frontiers.

\begin{algorithm}[!h]
\caption{OctoMap Update}
\label{alg:octomap_exploration}
\begin{algorithmic}[1]
\Require Submap point cloud $\mathcal{S} = \{s_1, s_2, ..., s_n\}$
\Require LiDAR origin point $\bar{s}$
\Require OctoMap tree $M$
\State Calculate $E_{before}$ \Comment{Only for enhancement}
\For{each point $s_i \in \mathcal{S}$}
    \State Generate OctoMap point $q_i  \leftarrow s_i$
    \State $M$.updateNode($q_i$, \texttt{true}) \Comment{Mark as occupied}
    \State $M$.insertRay($\bar{s}$, $q_i$) \Comment{Update free space along ray}
\EndFor
\State updateFreeSpace($M$) \Comment{Alg. \ref{alg:update_free_space}}
\State $\mathcal{I}$ = $E_{before} - E_{after}$ \Comment{Only for enhancement}
\State updateFrontiers($correction$=\texttt{true}) \Comment{Alg. \ref{alg:frontier_update}}

\end{algorithmic}
\end{algorithm}

\begin{algorithm}
\caption{Update Free Space with Forward Sensor Model}
\label{alg:update_free_space}
\begin{algorithmic}[1]
\Require OctoMap tree $M$
\Require LiDAR origin point $\bar{s}$
\Require Sensor Model $\mathcal{H}$
\State \textbf{function} updateFreeSpace()

\For{each ray $r$}    
    \State $hit \leftarrow M$.castRay($\bar{s}$, $r$)    
    \If{\textbf{not} $hit$} \Comment{No obstacle encountered}
        \State $\mathcal{K} \leftarrow M$.computeRayKeys($\bar{s}$, $r$)
        \If{$\mathcal{K}$ is valid}
            \For{each voxel key $k \in \mathcal{K}$}
                \State $node \leftarrow M$.search($k$)
                \If{$node = \texttt{null}$ \textbf{or} \textbf{not} occupied}
                    \State $M$.updateNode($k$, \texttt{false})
                    \State updateFrontiers(\texttt{false}) \Comment{Alg. \ref{alg:frontier_update}}
                \EndIf
            \EndFor
        \EndIf
    \EndIf
\EndFor

\State \textbf{end function}

\end{algorithmic}
\end{algorithm}

Along similar lines to the OctoMap voxel update method presented in  \cite{wurm2010octomap}, we adopt a beam-based inverse sensor model following the $3$D Bresenham algorithm  \cite{amanatides1987fast}. In this approach, a ray is cast from the sensor to the point, and the voxels along the ray are initialized with an occupancy probability as per Eq. \ref{eq:OccupancyProbabilityMinimal}. With each new measurement, the voxel probability is updated, informed by the prior observations. To quantify the information gain during an OctoMap update, we track the corresponding change in entropy. Since this evaluation is informed by actual ray-casting, it captures spatial realities much better than rigid geometric calculations. Let $M_{\mathrm{bbx}[\mathcal{S}]}$ denote the set of voxels in $M$ within the bounding box constraining the submap point cloud $\mathcal{S}$. The entropies $E_{before}$  and $E_{after}$ are calculated before and after the OctoMap update, respectively. The entropy is calculated by summing the binary entropy of each voxel $c \in M_{\mathrm{bbx}[\mathcal{S}]}$ based on its occupancy probability $p_{\mathrm{o}}(c)$:
\begin{equation*} \label{eq:Entropy}
-\sum_{c \in M_{\mathrm{bbx}[\mathcal{S}]}}p_{\mathrm{o}}(c) \log_2 p_{\mathrm{o}}(c) + (1-p_{\mathrm{o}}(c)) \log_2(1-p_{\mathrm{o}}(c)).
\end{equation*} The code flow for an OctoMap update is detailed in Alg. \ref{alg:octomap_exploration}

\subsubsection{Forward Detection and Maintenance}
An approximate frontier detection approach with the ray $r$ would classify a voxel as a frontier if it is free and the subsequent voxel along $r$ is unknown. We maintain a hashmap $\mathcal{F}$ of candidate frontiers. 
\begin{algorithm}
\caption{Adaptive Frontier Detection and Maintenance}
\label{alg:frontier_update}
\begin{algorithmic}[1]
\Require OctoMap tree $M$
\Require Correction flag $correction$
\Require LiDAR point $\bar{s}$, ray end point $p_{end}$
\Require Current frontier hashmap, $\mathcal{F}$

\State \textbf{function} updateFrontiers($correction$)

\If{$correction = \texttt{true}$} \Comment{Validate frontiers}
    
	\For{each frontier $f \in \mathcal{F}$} \Comment{Cost : $\mathcal{O}{(|\mathcal{F}|)}$}
        
        \If{$M$.search($f$) $\neq \texttt{null}$} 
            \State $\mathcal{F}$.erase($f$) \Comment{Delete Known Voxel}
            \State \textbf{continue}
        \EndIf
        
        \If{frontierWithinRange($f$,$\bar{s}$)} 
            \State $\mathcal{F}$.erase($f$) \Comment{Frontier too close to robot}
            \State \textbf{continue}
        \EndIf

    \EndFor
    
\Else \Comment{Add new frontier behind voxel}
    \State $\vec{r} \leftarrow (p_{end} - \bar{s})$.normalized()
    \State $p_{behind} \leftarrow p_{end} + \vec{r} \times M$.getResolution()
        \If{$M$.search($p_{behind}$) $= \texttt{null}$} 
            \State $\mathcal{F}$.add($p_{behind}$) 
    \EndIf
\EndIf

\State \textbf{end function}

\end{algorithmic}
\end{algorithm}

Consider a sensor at pose $\bar{s}$ providing a measurement $z$ of the surrounding environment. We define an approximate sensor model $\mathcal{H}$, which encodes characteristics such as range constraints $\lambda_{\max}$, fields of view $({\Phi}, {\Psi})$, and angular resolutions. Using $\mathcal{H}$, a ray $r$ originating from the sensor can be expressed as:
\begin{equation*} \label{eq:Raycasting}
r(\lambda; \phi, \psi) = \bar{s} + \lambda \hat{d}(\phi, \psi),
\end{equation*}
where $\hat{d}(\phi, \psi)$ is the unit vector in the direction $(\phi \in \Phi, \psi \in \Psi)$ relative to the sensor. Following a first-hit model, the estimated ray length $\hat{\lambda}$ propagated through the OctoMap $M$ from the origin $\bar{s}$ is given by:
\begin{equation} \label{eq:Rayestimation}
\hat{\lambda}(\phi, \psi) = \min \{ \lambda > 0 \mid p_{\mathrm{o}}(\underbar{m}(M, r(\lambda; \phi, \psi))) > \tau \},
\end{equation}
where $\underbar{m}(M, \mathbf{x})$ denotes the leaf node at position $\mathbf{x}$ and $\tau$ is the occupancy threshold. $p_{\mathrm{o}}(\underbar{m}(M, r(\lambda; \phi, \psi)))$ represents the occupancy probability of the voxel intersected by $r$ at distance $\lambda$ under the constraints imposed by the sensor model $\mathcal{H}$. The parameter $\tau$ serves as the occupancy threshold; thus, Eq. \ref{eq:Rayestimation} identifies the ray's endpoint as the first occupied voxel along the propagation path. This formulation enables efficient free-space updates, as detailed in Alg. \ref{alg:update_free_space}.

With each OctoMap update, frontier detection is integrated directly into the free space update procedure (Alg.~\ref{alg:update_free_space}). This involves casting a fixed number of rays determined by the sensor model $H$. Since the number of rays is constant for a given sensor, the work done to identify new frontier candidates per update is constant. The primary computational cost shifts purely to maintaining and validating the existing list of frontiers (Alg.~\ref{alg:frontier_update}, lines $3$--$11$), which requires iterating through the entire set $\mathcal{F}$, incurring a total complexity of $\mathcal{O}(|\mathcal{F}|)$. This is asymptotically lower than that of the most comparable OctoMap-based frontier detection approach~ \cite{batinovic2021multi}, which exhibits a complexity of $\mathcal{O}(\underline{c} + |\mathcal{F}|)$, where $\underline{c}$ denotes the number of changed or updated cells in the OctoMap. While $\mathcal{O}(\underline{c} + |\mathcal{F}|)$ successfully decouples frontier detection from the total map volume, it remains highly vulnerable to severe computational spikes when the robot observes massive new volumes of space simultaneously (i.e., when $\underline{c}$ explodes). By integrating frontier detection directly into the forward sensor model operating on a fixed number of rays, our approach completely eliminates the dependency on $\underline{c}$, ensuring a strictly smoother, predictable computational profile.

\subsection{Clustering}
To reduce redundancy and form compact exploration targets, the detected frontiers $\mathcal{F}$ are clustered. Mean-shift clustering  \cite{fukunaga1975estimation} is adopted due to its non-parametric nature, avoiding the need to predefine the number of clusters. Following  \cite{batinovic2021multi}, a Gaussian kernel is used, but with fixed parameters for robustness across environments. The resulting cluster centers serve as the candidate viewpoints, with each candidate denoted as $v_c$.

\subsection{Best Viewpoint Selection}\label{subsubsec:bestviewpointselection}
To directly compare the proposed approximate frontier detection to \cite{batinovic2021multi}, we adopt the same  viewpoint prioritization method. The exploratory gain at each candidate viewpoint, $v_c$, for a robot position $\bar{\mathcal{R}}$ is given by

\begin{equation} \label{eq:ExplorationGain}
G(v_c) = \frac{I(v_c)}{e^{-\lambda_g\|{\bar{\mathcal{R}}-{v_c}}\|}},
\end{equation}

where $I(v_c)$ is the deterministic information gain, the measure of the unexplored area that can be visualized from $v_c$. $I(v_c)$ is approximated as the number of unknown voxels within a cube centered at $v_c$. The constant $\lambda_{g}$ tunes the choice between distance vs exploratory gain. The ideal viewpoint that maximizes $G(v_c)$ is then selected as the best viewpoint.

\begin{figure*}[!htb]
    \centering
    \begin{subfigure}{0.18\textwidth}
        \centering
        \includegraphics[width=\textwidth, height=0.8\textwidth]{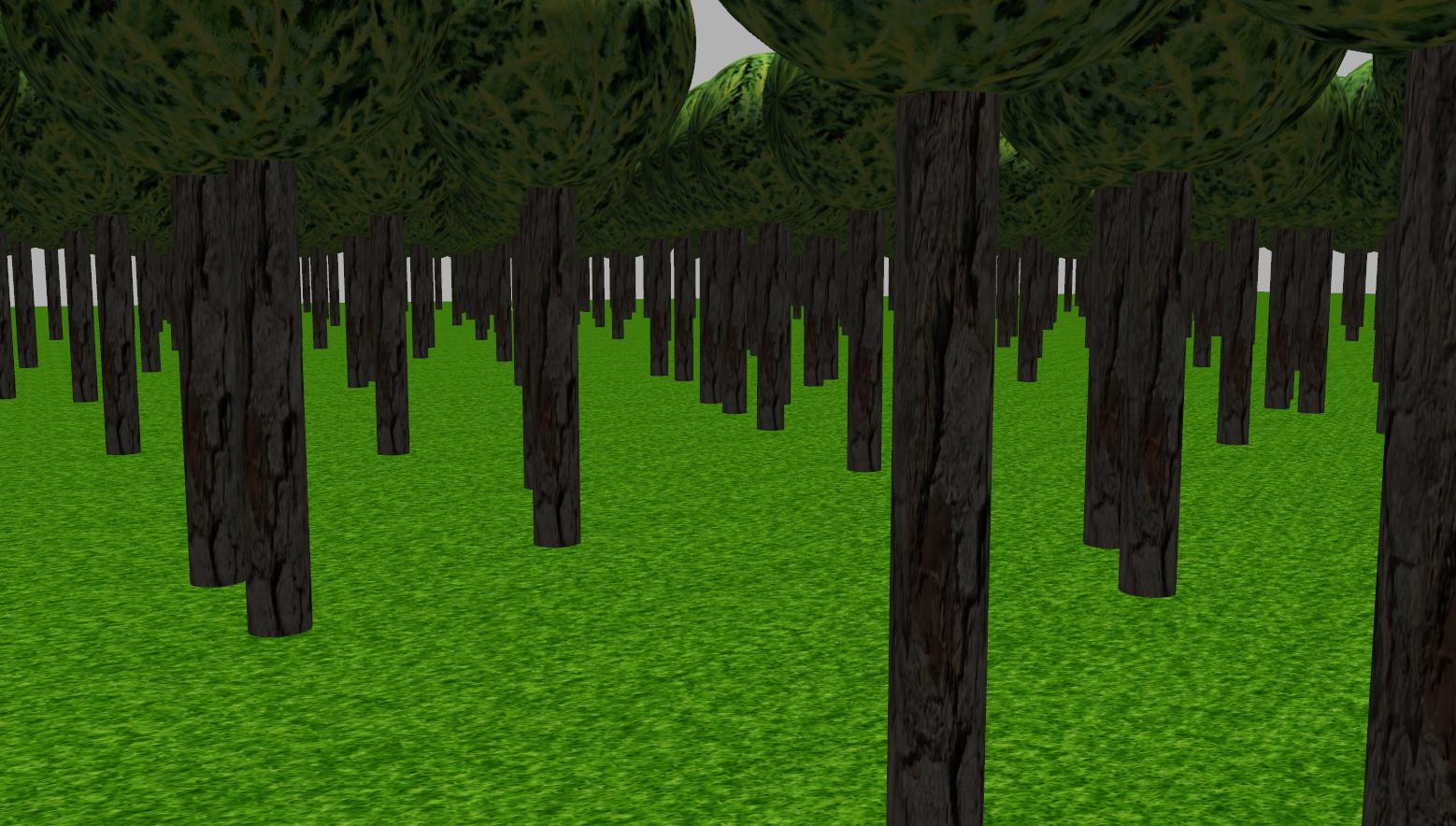}
        \caption{Forest}
        \label{fig:env_forest}
    \end{subfigure}
    \hfill
    \begin{subfigure}{0.18\textwidth}
        \centering
        \includegraphics[width=\textwidth, height=0.8\textwidth]{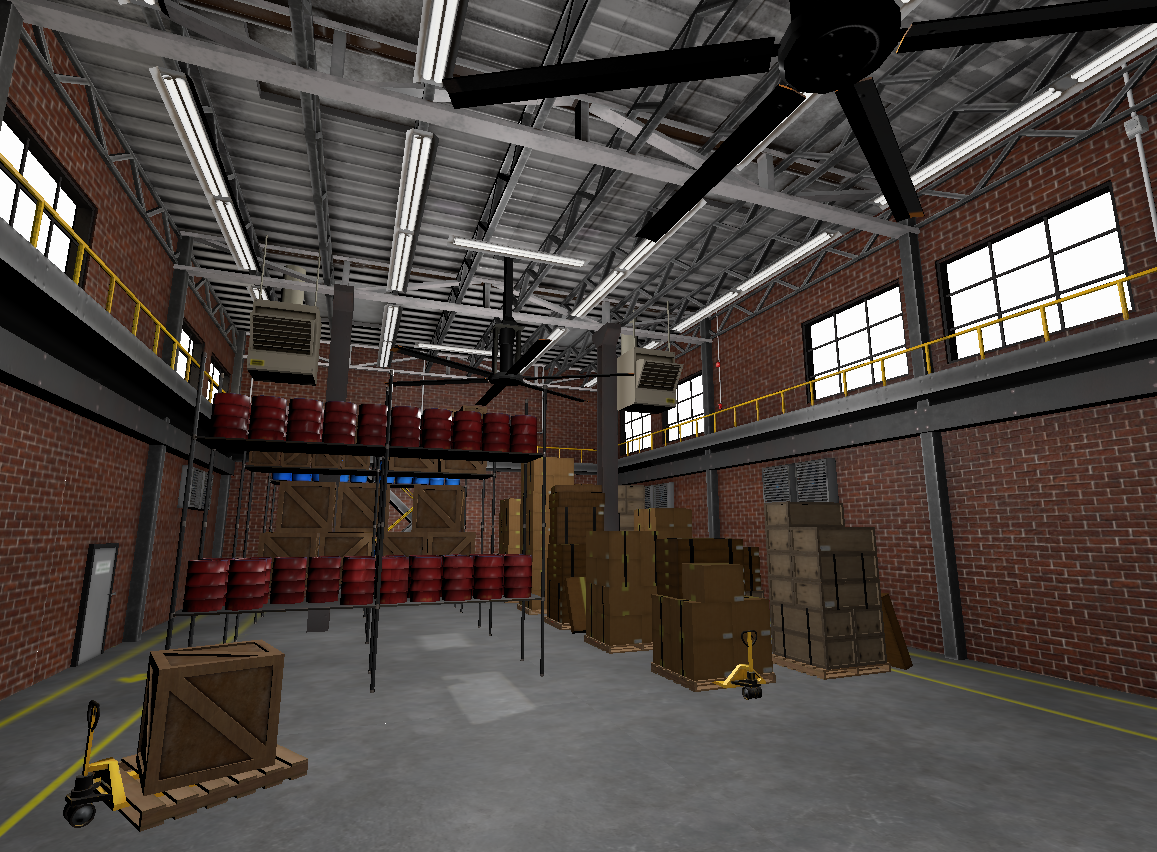}
        \caption{Warehouse}
        \label{fig:env_warehouse}
    \end{subfigure}
    \hfill
    \begin{subfigure}{0.18\textwidth}
        \centering
        \includegraphics[width=\textwidth, height=0.8\textwidth]{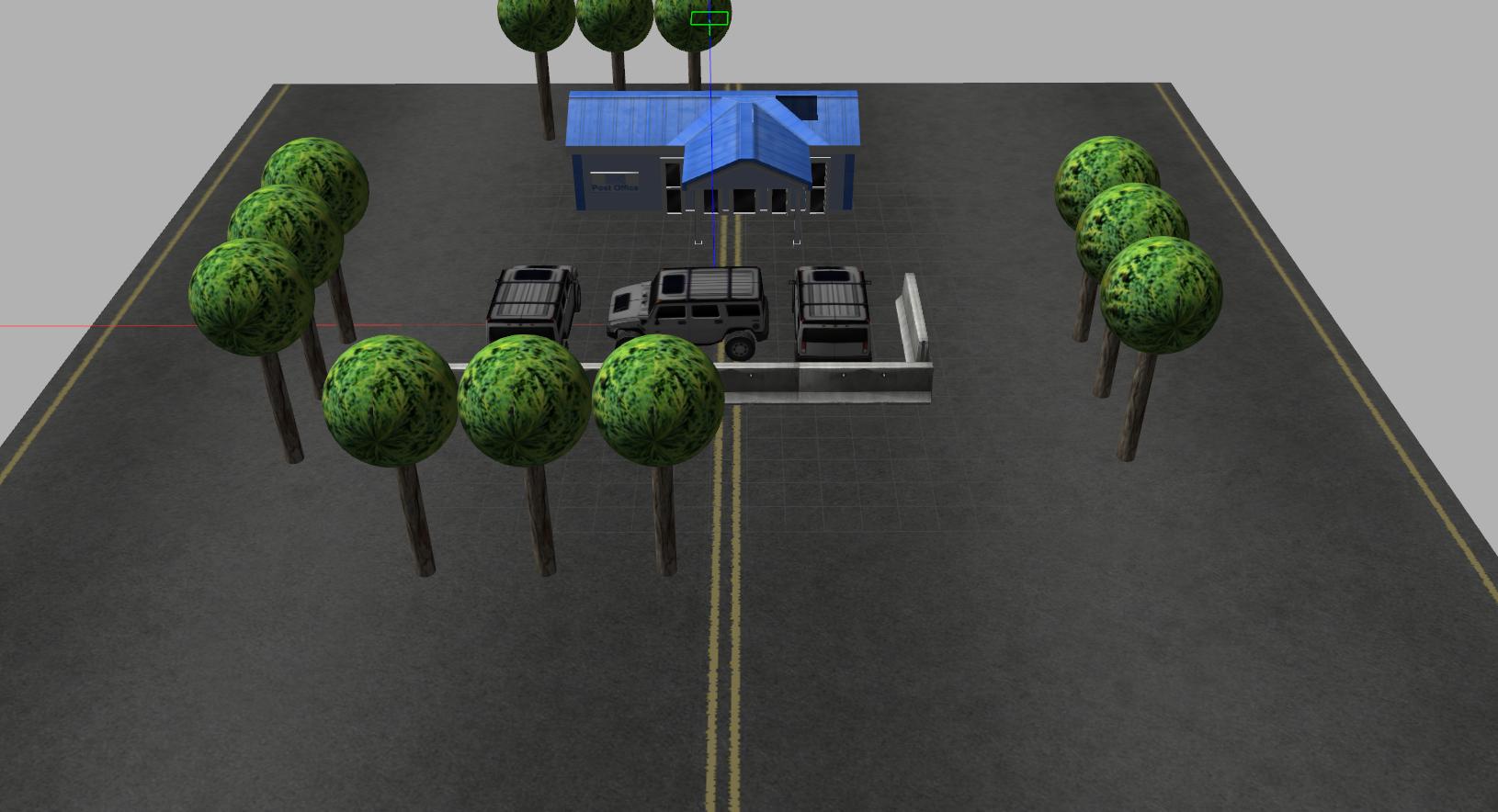}
        \caption{House Compound}
        \label{fig:env_house}
    \end{subfigure}
    \hfill
    \begin{subfigure}{0.18\textwidth}
        \centering
        \includegraphics[width=\textwidth, height=0.8\textwidth]{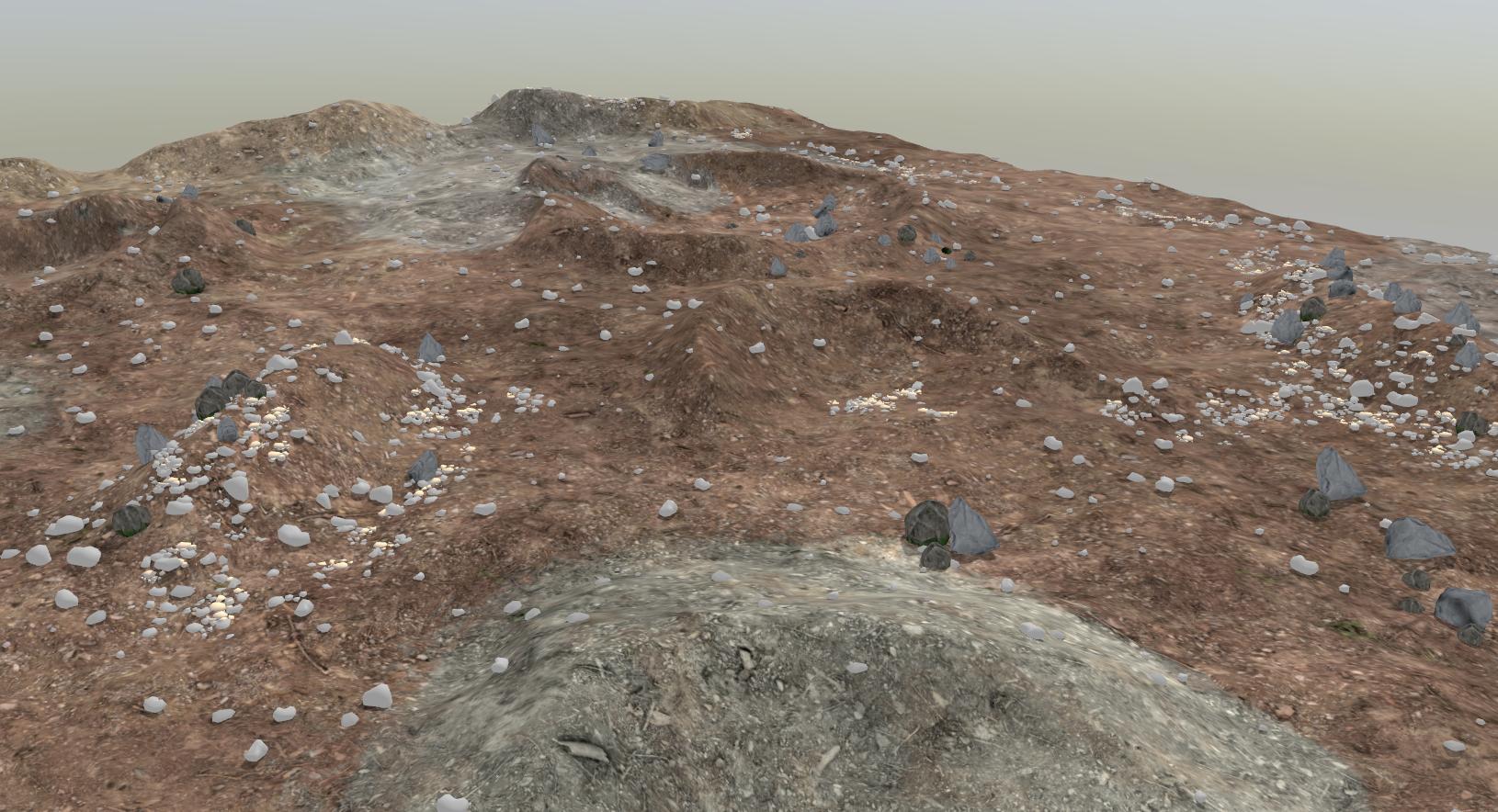}
        \caption{Mars Surface}
        \label{fig:env_mars}
    \end{subfigure}
    \hfill
    \begin{subfigure}{0.18\textwidth}
        \centering
        \includegraphics[width=\textwidth, height=0.8\textwidth]{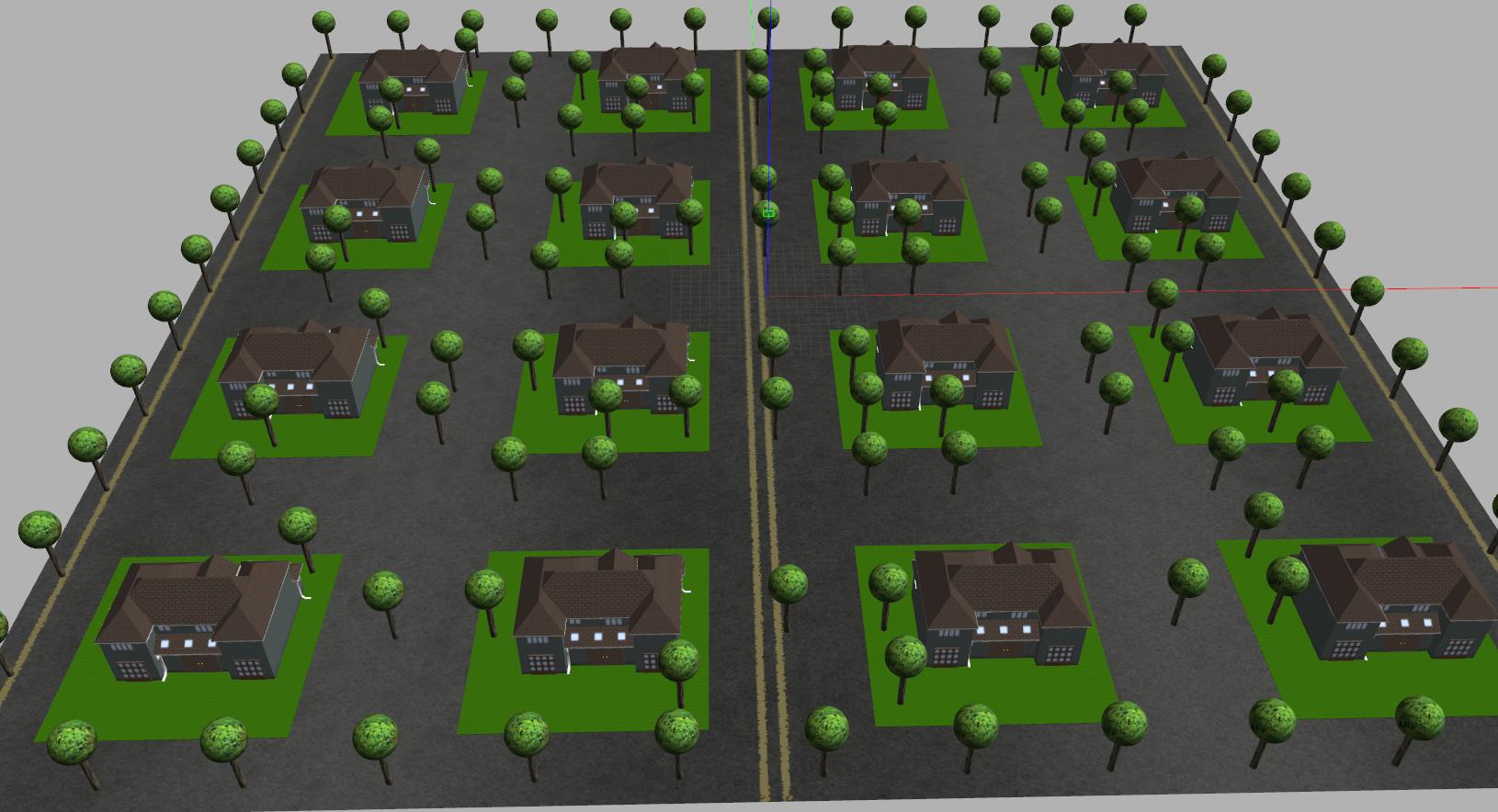}
        \caption{Housing Colony}
        \label{fig:env_housing_colony}
    \end{subfigure}    
    \caption{Simulation environments used for analysis. 
    {\textbf{Forest \cite{baca2021mrs}:}} Unstructured cluttered environment with a tree density of 0.05 trees/$m^2$,
    {\textbf{Warehouse:}} Cluttered indoor environment, 
    {\textbf{House Compound:}} Structured environment, 
    {\textbf{Mars Surface:}} Minimal feature uneven ground environment,
    {\textbf{Housing Colony:}} Structured symmetric environment.     
    }
    \label{fig:Results-SimulationEnvironments}
\end{figure*}

\subsection{Path Planning}
Once the best viewpoint is determined, it is then passed to an OctoMap-based path planner for navigating to the viewpoint. A new viewpoint is sent to the path planner, when the robot is within a threshold distance from the previously selected viewpoint.

\subsection{\textbf{Enhancement:} Bayesian Regressor}
The deterministic information gain calculation method described in Eq.  \ref{eq:ExplorationGain} is computationally efficient but can be inaccurate, as it relies on a rigid geometric approximation of the unknown space around the candidate viewpoint. This can lead to suboptimal viewpoint selection, especially in complex environments where the spatial distribution of unknown voxels is not uniform. To address this limitation, we propose an enhancement that formulates the information gain estimation as a probabilistic regression problem using a Gaussian Process (GP) regressor.

To model the calculated information gain ($\mathcal{I}$ from Alg.\ref{alg:octomap_exploration}) using a Gaussian Process (GP), we define the probabilistically estimated information gain, $\mathcal{I}'$, as a function of the voxel feature vector $\mathbf{x} \in \mathbf{R}^3$, representing the raw counts of free, occupied, and unknown cells. To ensure robust length-scale scaling across varying map sizes, the raw feature vectors are dynamically normalized to a unit hypercube $\tilde{\mathbf{x}} \in [0, 1]^3$ based on the minimum ($\mathbf{x}_{\min}$) and maximum ($\mathbf{x}_{\max}$) bounds of the current data window
\begin{equation*}
\tilde{\mathbf{x}} = \frac{\mathbf{x} - \mathbf{x}_{\min}}{\mathbf{x}_{\max} - \mathbf{x}_{\min}}
\end{equation*}
where the division is applied element-wise. Assuming a prior mean based on the empirical data average $\mu$, the model is formulated as:
\begin{equation*}
\mathcal{I}'(\tilde{\mathbf{x}}) \sim \mathcal{GP}(\mu, k(\tilde{\mathbf{x}}, \tilde{\mathbf{x}}')),
\end{equation*}
where the spatial correlation between any two viewpoints in the normalized feature space is captured using a Squared Exponential kernel \cite{rasmussen2003gaussian}, given by:
\begin{equation*}
k(\tilde{\mathbf{x}}, \tilde{\mathbf{x}}') = \sigma_f^2 \exp\left(-\frac{\|\tilde{\mathbf{x}} - \tilde{\mathbf{x}}'\|^2}{2l^2}\right).
\end{equation*}
Here, $\sigma_f^2$ represents the signal variance and $l$ denotes the length-scale parameter, which dictates the smoothness of the predicted information gain across the feature space.

With every new update of the observed $\mathcal{I}$ and the associated viewpoint features, the model bounds are recalculated, and the GP is updated using a sliding window of the $W$ most recent viewpoint evaluations to maintain computational tractability. Let $X$ and $\mathbf{y}$ represent the normalized feature vectors and observed information gains in the current window, and $K$ be their covariance matrix. To estimate $\mathcal{I}'$ for a newly generated candidate viewpoint $v_c$, its corresponding feature vector $\mathbf{x}_c$ is first normalized to $\tilde{\mathbf{x}}_c$, and its correlation vector $\mathbf{k}_c$ is evaluated against the existing knowledge base to compute the posterior predictive mean \cite{rasmussen2003gaussian}:
\begin{equation*}
\mathcal{I}'(\tilde{\mathbf{x}}_c) = \mu + \mathbf{k}_c^T (K + \sigma_n^2 I)^{-1} (\mathbf{y} - \mu \mathbf{1}),
\end{equation*}
where $\mathbf{k}_c$ is the correlation vector between the new viewpoint and the existing viewpoints, $\sigma_n^2$ is the observation noise variance and $\mathbf{1}$ is a vector of ones. 

While the covariance matrix inversion inherently carries an $\mathcal{O}(W^3)$ complexity, capping the sliding window at a fixed scalar restricts this to a constant-time factor. More importantly, estimating the utility for a set of $C$ candidate viewpoints requires only $\mathcal{O}(C \cdot W)$ operations, as the inverse covariance term is pre-computed during the update. This guarantees consistent real-time performance regardless of the global map size. In comparison with traditional deterministic approaches similar to Sec. \ref{subsubsec:bestviewpointselection}, the probabilistic approach enables structural generalization based on volumetric density. In large-scale, unstructured environments like forests, complex semantic recognition is unnecessary for effective exploration. The raw voxel distribution implicitly encodes the localized density signature (e.g., highly cluttered tree canopies versus open clearings). By mapping these local voxel distributions to expected information gains, the Gaussian Process learns the exploratory utility of distinct volumetric density profiles. This allows the system to rapidly prioritize open frontiers over dead-end clutter without relying on explicit geometric or semantic feature extraction. Finally, the adaptability and performance of this regressor are governed by the tunable hyperparameters $l$, $\sigma_f^2$, $\sigma_n^2$, and $W$, which allow the system to effectively balance predictive smoothness, observation confidence, and computational load.
\section{EXPERIMENTS \& RESULTS} 
\label{sec:results}
\subsection{Simulation} \label{Results:Simulation}
\subsubsection{Setup}\label{Results:Simulation_Setup}

\begin{table*}[h]
\centering
\renewcommand{\arraystretch}{1.3}
\resizebox{\textwidth}{!}{
\begin{tabular}{lcccccccc}
\toprule
Env & Forest & Forest & Forest & Forest & Warehouse & House Compound & Mars Surface & Housing Colony \\
Alg/Volume & $1200 m^3$ & $2700 m^3$ & $4800 m^3$ & $7500 m^3$ & $1260 m^3$ & $2688 m^3$ & $6336 m^3$ & $9408 m^3$ \\
\midrule
Hybrid\cite{ribeiro2024efficient} & \textbf{43.87 $\pm$ 12.98} & \underline{$74.35 \pm 24.86$} & \underline{$117.76 \pm 21.15$} & \underline{$178.18 \pm 26.88$} & $42.95 \pm 54.86$ & $192.31 \pm 74.38$ & $186.27 \pm 55.63$ & \underline{$476.82 \pm 78.73$} \\
EPIC\cite{EPIC} & $47.66 \pm 22.24$ & $110.28 \pm 15.29$ & $147.26 \pm 14.67$ & $224.06 \pm 25.36$ & \textbf{41.42 $\pm$ 8.14} & $135.93 \pm 16.83$ & \textbf{119.85 $\pm$ 40.74} & $-$ \\
Frontier\cite{batinovic2021multi} & $66.74 \pm 14.37$ & $125.65 \pm 31.33$ & $216.70 \pm 25.88$ & $303.39 \pm 32.10$ & $61.34 \pm 12.31$ & \textbf{110.56 $\pm$ 18.80} & $213.16 \pm 53.94$ & $582.35 \pm 89.52$ \\
Asymp(p) & $55.18 \pm 9.59$ & $80.75 \pm 23.90$ & $137.19 \pm 22.04$ & $227.44 \pm 28.87$ & $47.55 \pm 6.60$ & $137.88 \pm 22.75$ & $173.39 \pm 28.03$ & $544.26 \pm 48.61$ \\
Asymp+Bayes(p) & \underline{$45.09 \pm 7.10$} & \textbf{57.24 $\pm$ 13.06} & \textbf{106.14 $\pm$ 16.91} & \textbf{177.20 $\pm$ 24.89} & \underline{$42.76 \pm 6.11$} & \underline{$112.05 \pm 18.79$} & \underline{$143.82 \pm 36.04$} & \textbf{444.73 $\pm$ 17.98} \\
\bottomrule
\end{tabular}
}
\caption{Exploration duration statistics(seconds) across all environments. \textbf{Emboldened} values indicate the method with the fastest exploration for a given environment, while the \underline{underscored} depict the second fastest.}
\label{tab:merged_exploration_results}
\end{table*}

\begin{figure*}[ht]
  \centering
  \begin{minipage}[h]{\textwidth}
      \centering
      \includegraphics[width=\textwidth]{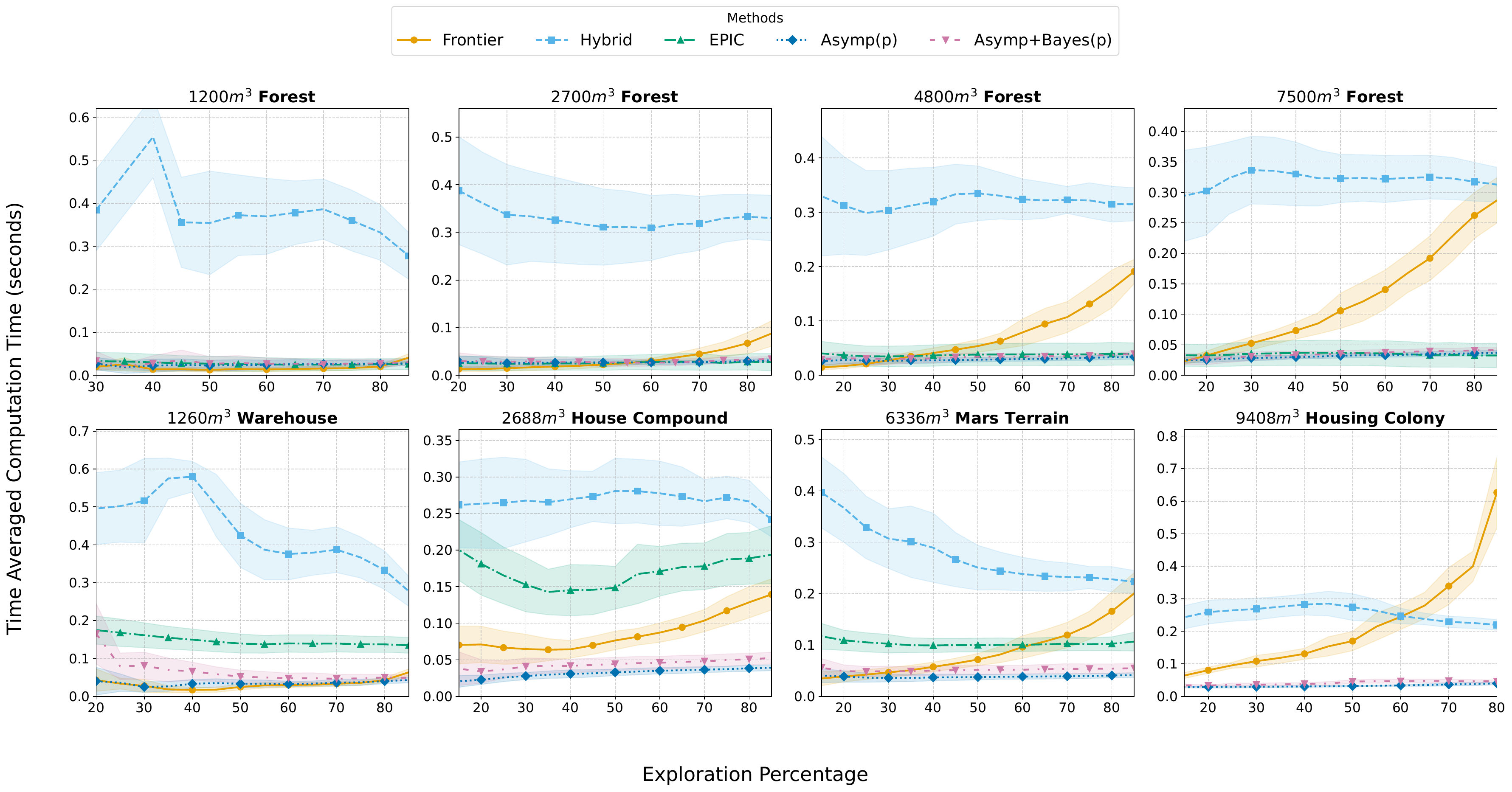}
     \caption{Time Averaged Computation across coverage percentage for varying methods for each environment.}
      \label{fig:Results-computation-time}
	\end{minipage}
\end{figure*}

Simulations are carried out in the MRS-UAV framework \cite{baca2021mrs} running on the Gazebo simulator with ROS Noetic. The choice was motivated by the simulator's realistic Software-In-The-Loop (SITL) simulations and ease of simulation-to-real-world transition. The maximum (horizontal speed, vertical speed, acceleration, jerk) were set to ($1$m/s, $1$m/s, $1$m/s$^2$, $20$m/s$^3$). The $3D$ perception sensor of choice is a downward-facing Mid-360 Livox on a f$550$ UAV. The sensor pointcloud, range limited to $20$m, is fed into a SLAM module \cite{reinke2022locus} to generate the $3D$ pointcloud representation, the submap of which generates the OctoMap \cite{wurm2010octomap} for path planning \cite{kratky2021autonomous}. 

To benchmark the proposed method in both computational expense and exploration time, we compare the algorithm against other OctoMap-based exploration methods like  \cite{batinovic2021multi}, which is a frontier-based exploration method aimed at relieving computational requirements. For an improved analysis, we compare the method against the hybrid OctoMap-based exploration method \cite{ribeiro2024efficient} and the lightweight LiDAR explorer, EPIC \cite{EPIC}. It is worth noting that EPIC has recently been demonstrated to outperform other state-of-the-art map-agnostic frameworks, such as  \cite{zhou2021fuel,10582913}. Therefore, benchmarking against EPIC provides a rigorous assessment against a highly competitive, high-performance frontier exploration baseline, validating our method's standing relative to current state-of-the-art architectures. For ease of depiction, we will be referring to  \cite{batinovic2021multi} as \textit{Frontier},  \cite{ribeiro2024efficient} as \textit{Hybrid} and  \cite{EPIC} as \textit{EPIC} in our results. Similarly, the proposed approach without the Bayesian regressor will be referred to as \textit{Asymp(p)}, and with the Bayesian regressor enhancement as \textit{Asymp+Bayes(p)}.

With exploration time capped at $1000$ secs, the experiments were carried out on an Ubuntu-$20$ Apptainer run on an HPC allocated with $32$GB RAM. The averaged results across $100$ runs are then analyzed. The simulation environments were chosen for their unique features as described in Fig.\ref{fig:Results-SimulationEnvironments}. An exploration task is considered to be successful if $90\%$ of the environment is perceived by the UAV. For all environments, the maximum flight altitude is set to $3$m. To analyze the effect of a constant environment with varying sizes, the exploration methods are tested in the forest environment for varying sizes. Due to the complete occlusion of house interiors in the housing colony (Fig.\ref{fig:env_housing_colony}), the success percentage reduced to $85\%$. 

The resolution of the respective 3D representation is set to $0.4$ across all methods, and each method's parameters are tuned to improve performance in the environment. The Bayesian regressor is implemented using  \cite{LIMBO}. To ensure real-time performance during autonomous exploration, the sliding window size for the Bayesian Regressor was fixed at $W=200$. The hyperparameters $(\sigma_f^2,l,\sigma_n^2)$ are set to $(1.0,0.3,0.01)$ to ensure a smooth predictive generalization across the feature space, and minor heuristic approximations.

The primary figures of merit used for our evaluation are the total time for successful exploration and the time averaged computation time required for every $5\%$ increment of volume coverage. For a given exploration method, the total computation time is defined as the sum of the times required to: 1) update the map representation, 2) frontier detection 3) the next-best viewpoint determination, and 4) generate a safe trajectory to the selected viewpoint. Since exploration times naturally vary across different methods, we normalize these metrics by averaging the computation time over the duration required to reach each $5\%$ milestone to ensure a consistent baseline for direct comparison. A detailed breakdown of the modules is showcased in the attached multimedia.


\subsubsection{Computation Time}
Fig.\ref{fig:Results-computation-time} presents the time-averaged computation time across coverage percentage for varying methods in each environment. The results highlight two key takeaways: 1) the proposed method's asymptotic computational efficiency across all environments, including scale invariance in the forest environments, and 2) despite the $\mathcal{O}(W^3)$ complexity of the covariance matrix inversion, setting $W=200$ restricts the computational overhead to a negligible constant factor. 

As empirically demonstrated in Fig.\ref{fig:Results-computation-time}, the total time-averaged computation for \textit{Asymp+Bayes(p)} closely tracks that of the deterministic approach, \textit{Asymp(p)}. This proves that the system reaps the exploratory speed benefits of probabilistic information gain estimation without suffering a debilitating computational penalty. Furthermore, when compared to other OctoMap-based approaches like \textit{Frontier} \cite{batinovic2021multi} and \textit{Hybrid} \cite{ribeiro2024efficient}, both proposed methods achieve consistently lower computation times. They even remain competitive with—and in structured environments, sometimes faster than—EPIC \cite{EPIC}, a method that completely bypasses the need to update and maintain an OctoMap. 

\subsubsection{Exploration Time}
Tab.\ref{tab:merged_exploration_results} showcases the exploration time statistics for successful runs analyzed across $100$ trials. The results highlight two primary takeaways: the baseline improvement of \textit{Asymp(p)} over the direct deterministic frontier-based method \cite{batinovic2021multi}, and the consistent, superior performance of \textit{Asymp+Bayes(p)} across all tested environments.

\textit{Asymp(p)} and the baseline \textit{Frontier} \cite{batinovic2021multi} share the same deterministic viewpoint selection approach, allowing a direct comparison of the quality of the frontiers they generate. \textit{Asymp(p)} demonstrates a clear improvement over the baseline, particularly highlighting scale invariance in the forest environments of varying sizes, yielding an average improvement of $28.7\%$. The only exception to this trend was observed in the House Compound environment, where \textit{Frontier} performed slightly better. This is largely due to the highly structured nature of the compound, which naturally mitigates the need for tight trajectory generation.

\begin{figure*}[ht]
    \centering
    \begin{minipage}{0.24\textwidth}
        \centering
        \includegraphics[height=0.75\textwidth, width=\linewidth ]{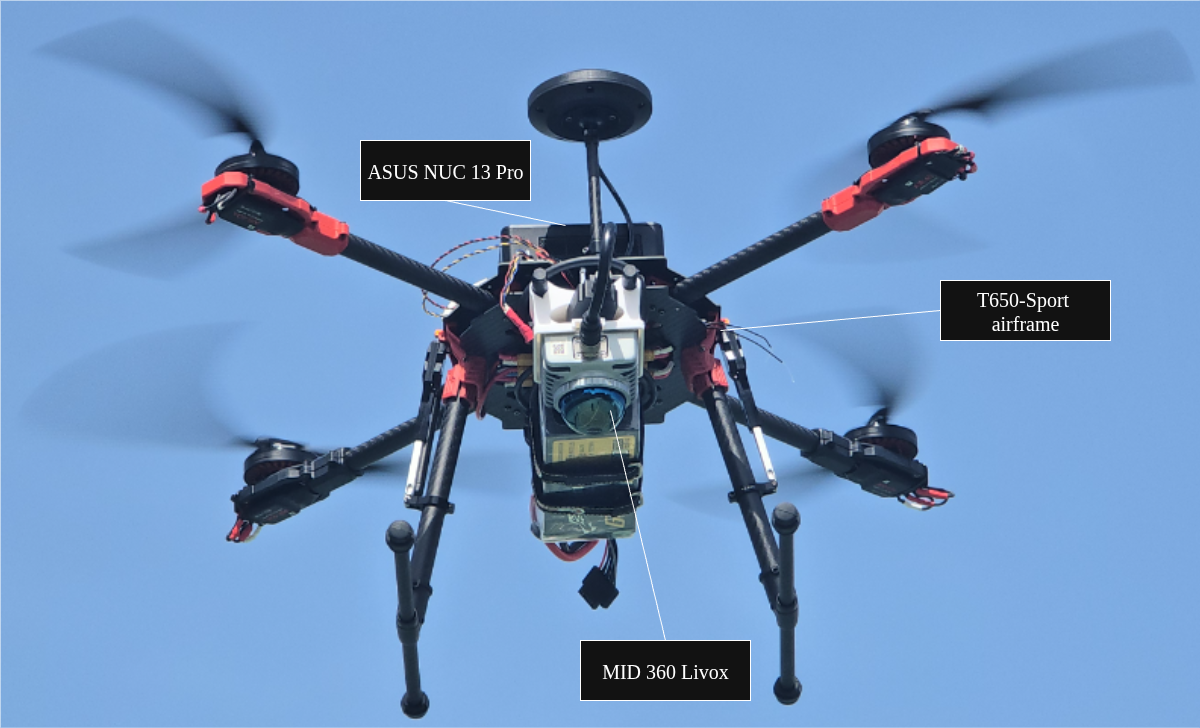}
        
        a)
    \end{minipage}
    \begin{minipage}{0.24\textwidth}
        \centering
        \includegraphics[height=0.75\textwidth, width=\linewidth, keepaspectratio]{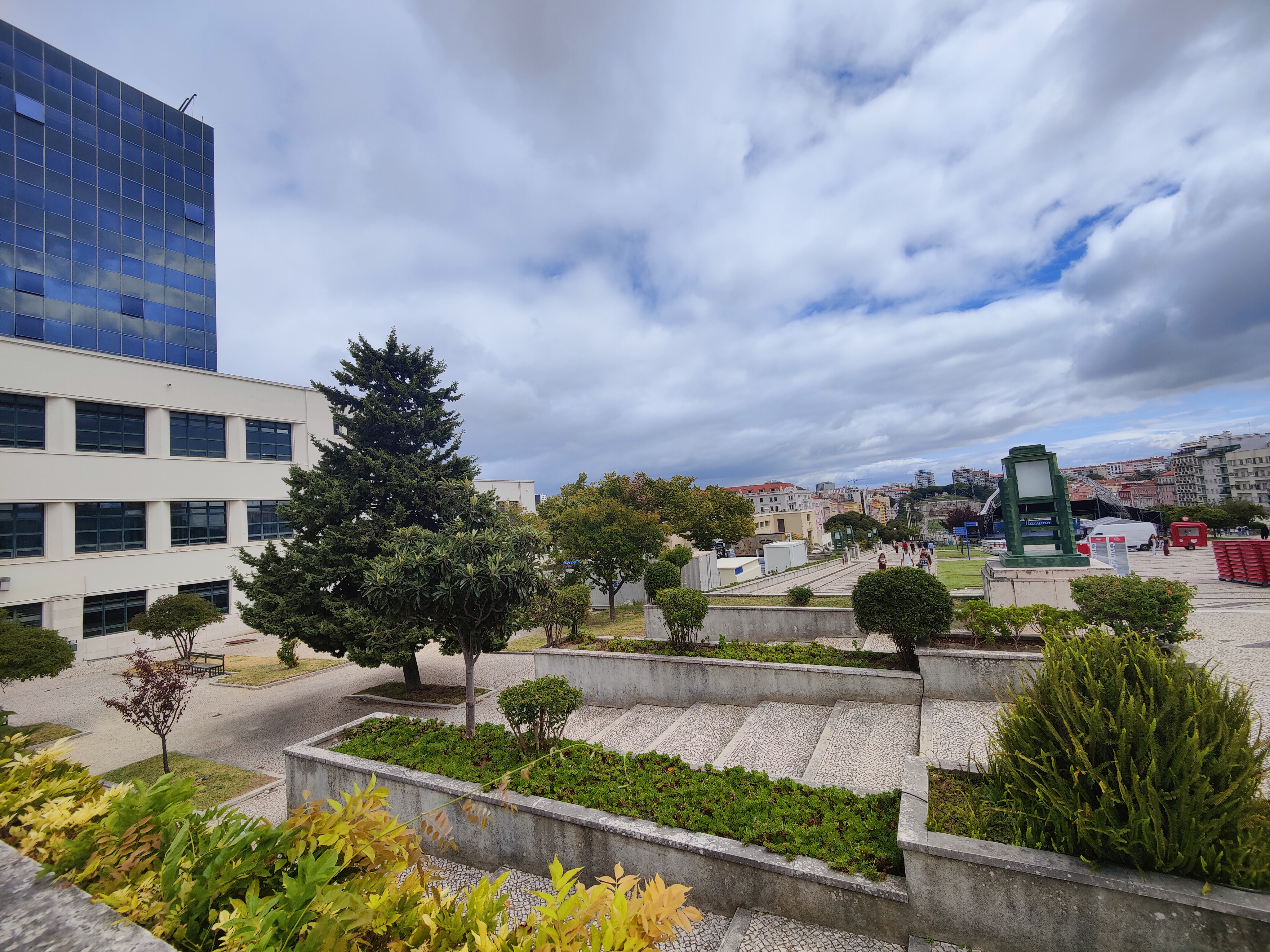}
        
        b)
    \end{minipage}
    \begin{minipage}{0.24\textwidth}
        \centering
        \includegraphics[height=0.75\textwidth, width=\linewidth]{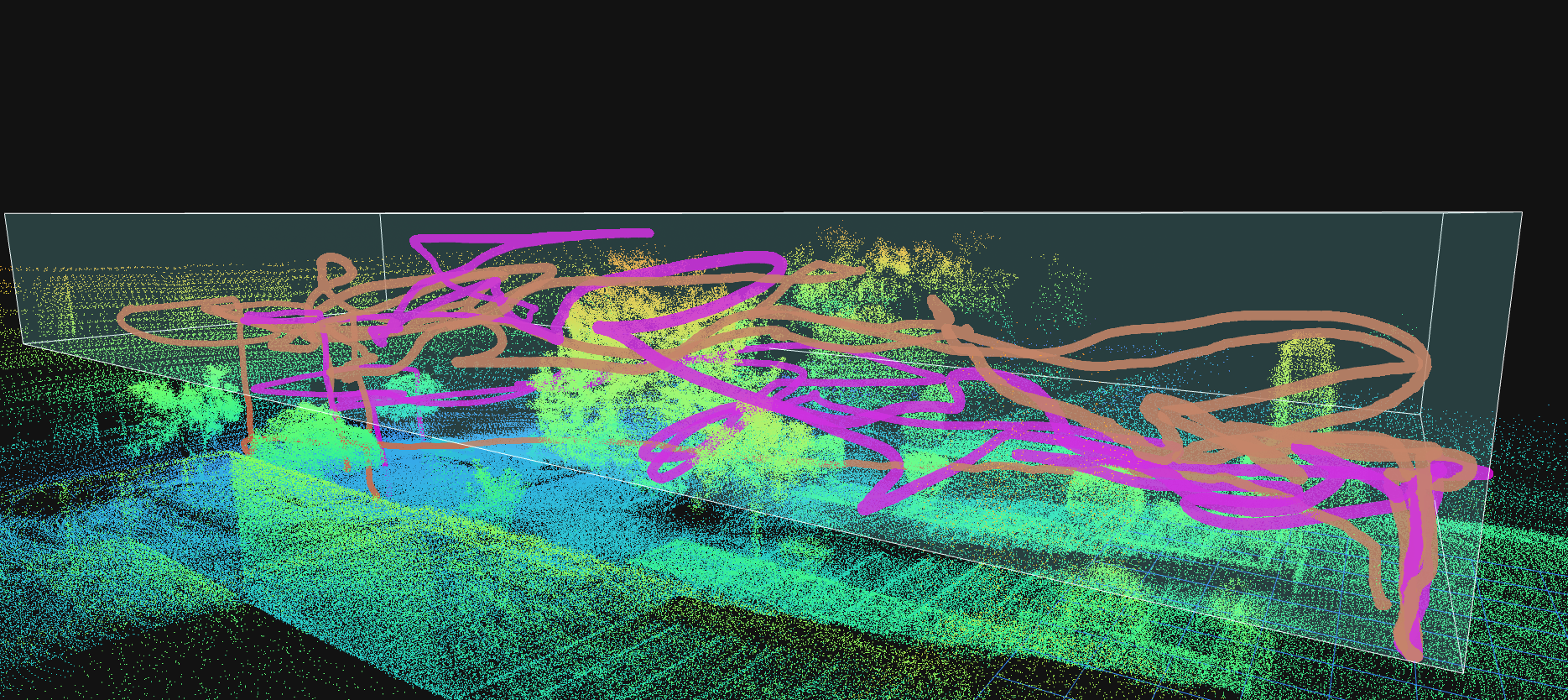}
        
        c)
    \end{minipage}
        \begin{minipage}{0.24\textwidth}
        \centering
        \includegraphics[height=0.75\textwidth, width=\linewidth]{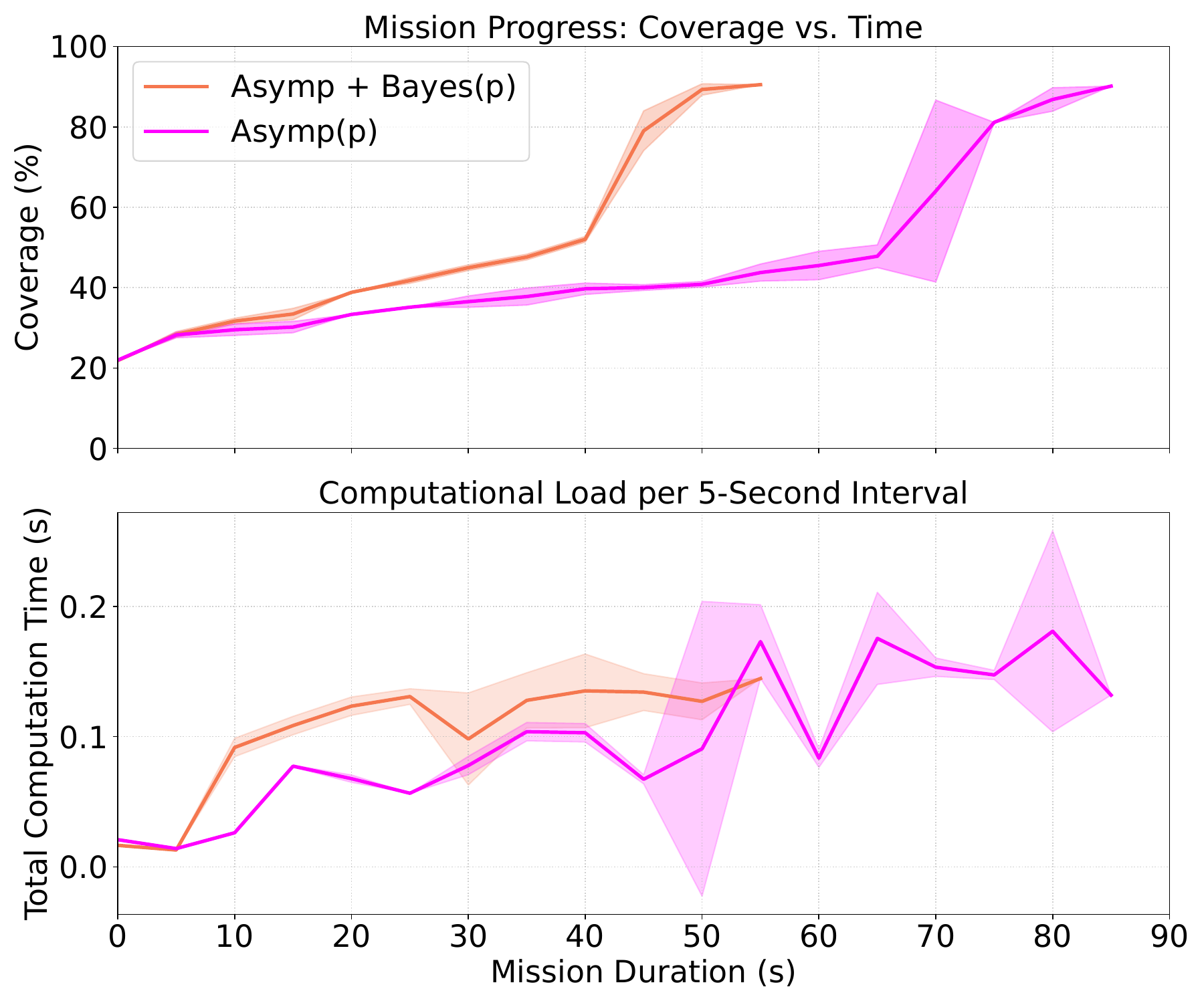}
        
        d)
    \end{minipage}
    \caption{Real-world setup. From left to right: \textbf{a)} T650-sport UAV equipped with Mid-360 Livox and ASUS NUC 13 Pro. \textbf{b)} Real-world Test Site. \textbf{c)} $3D$ SLAM Map of Real-World Environment with overlaying (\textcolor[HTML]{ff00ff}{Asymptotic}, \textcolor[HTML]{f5774f}{Asymp + Bayes}) paths and bounding cuboid. \textbf{d)} Trial averaged mean-std plots of coverage percentage and computation time vs Exploration time.}
    \label{fig:Results-Real_World}
\end{figure*}

Building on the base method, enhancing the approximate asymptotically-bounded frontier detection with a Bayesian regressor (\textit{Asymp+Bayes(p)}) further accelerates exploration. This improvement is driven by the regressor's ability to effectively estimate future information gain, allowing for smarter decision-making in selecting frontiers. It achieves a maximum exploration time improvement of $54.4\%$ over the \textit{Frontier} baseline in the $2700~m^3$ forest. When compared against the best baselines in the Forest environments, \textit{Asymp+Bayes(p)} achieved exploration time improvements of $23\%$, $9.9\%$, and $0.55\%$ for the $2700~m^3$, $4800~m^3$, and $7500~m^3$ spaces, respectively. 

While other methods occasionally excel in specific niches—such as EPIC \cite{EPIC} in small, structured environments or the clutter-less Mars surface—they lack generalization across varying complexities. Notably, despite multiple parameter tuning attempts, EPIC failed to achieve successful exploration in the Colony environment. This failure can be attributed to the symmetric nature of the environment combined with large occluded house interiors, which resulted in unreliable frontier generation and selection. Ultimately, by consistently minimizing exploration time across diverse scales and structures, \textit{Asymp+Bayes(p)} stands out as the most robust exploration strategy among the evaluated methods.

\subsection{Real World Experiments}

The real-world experiments are carried out in a T650-sport equipped with Mid-$360$ Livox and an ASUS NUC $13$ Pro for computation (Fig.\ref{fig:Results-Real_World}.a). The framework is deployed on an Apptainer running ROS Noetic powered by Ubuntu 20 on Intel Core i$7-1360$P CPU @ $2.2$GHz$\times 16$. The maximum (velocity, acceleration) of the UAV is limited to ($1m/s$, $1m/s^2$). While the simulated environments validate the algorithm's performance at scale, the real-world experiments are designed to validate the framework's computational tractability on resource-constrained hardware. The UAV is tasked to explore a bounding cuboid ($36$m$\times12$m$\times4$m) in the environment shown in Fig.\ref{fig:Results-Real_World}.b. Running the exploration framework on an onboard CPU (Intel Core i7-1360P)  in a challenging, tight environment highlights its ability to rapidly generate safe, approximate frontiers without violating strict real-time control constraints. Fig.\ref{fig:Results-Real_World}.c overlays the SLAM map on the bounding cuboid along with the path taken using the proposed method, \textit{Asymp(p)} and the enhancement, \textit{Asymp+Bayes(p)} across the $6$ trials ($3$ each) conducted. The sub-map from SLAM is limited to $10$m. 

The accompanying video demonstrates SLAM and OctoMap updates alongside best-viewpoint selection, while showcasing the coverage and total computation time for a given run. Fig.\ref{fig:Results-Real_World} d) presents averaged results over $3$ runs, with datapoints approximated to $5$ second buckets of duration for ease of statistical analysis. Both methods show a consistent increase in coverage percentage, with \textit{Asymp+Bayes(p)} showcasing a faster increase in coverage percentage compared to \textit{Asymp(p)}. Furthermore, the total computation time for every $5$ second bucket is bounded well below $0.20$ seconds, with \textit{Asymp+Bayes(p)} closely tracking the computation time of \textit{Asymp(p)}. The real-world experiments validate the proposed method's ability to operate efficiently in complex, real-world environments.  

A direct, fair comparison with other baselines in real-world scenarios is challenging due to significant differences in experimental setups, including sensor types and environmental complexity. For example, the hybrid method  \cite{ribeiro2024efficient} used in simulation lacks real-world results, while the approach in  \cite{batinovic2021multi} was evaluated in open environments where a lower OctoMap resolution for frontier processing was sufficient. While  \cite{EPIC} demonstrates successful validation in complex real-world scenarios, head-to-head benchmarking remains restricted to simulation, highlighting the broader difficulty of standardizing hardware constraints, safety protocols, and sensor configurations across distinct exploration architectures.

\section{CONCLUSION}
\label{sec:conclusion}
In this article, we presented a computationally bounded frontier exploration algorithm, enhanced by a Bayesian regressor for information gain estimation. By integrating a forward sensor model with OctoMap's inverse sensor modeling, our approach efficiently detects frontiers while minimizing octree traversals and recursive updates—overcoming a common bottleneck in traditional exploration architectures. Furthermore, training the Bayesian regressor in parallel with the OctoMap updates allows for accurate information gain estimation without incurring additional computational overhead, enabling informed, real-time decision-making.

Comprehensive simulation analyses across diverse environments demonstrate our algorithm's superior computational efficiency without sacrificing exploration performance. Notably, by strictly decoupling the computational time from the global map size, our method proves highly advantageous for large-scale deployments. The integration of the Bayesian regressor further accelerates missions, achieving up to a $54\%$ improvement in exploration time over standard deterministic OctoMap baselines, while remaining highly competitive with, and frequently outperforming, state-of-the-art hybrid and map-agnostic approaches. While certain baselines maintain an advantage in highly specific environments—indicating areas for further optimization—the proposed framework offers robust, scalable autonomy.

Future work will extend the proposed algorithm to multi-robot systems, leveraging its efficient frontier detection to identify globally optimal exploration targets. Ongoing efforts are exploring its integration into active SLAM frameworks, utilizing frontier uncertainties to inform next-best-pose decisions. A planned implementation with UFOMap\cite{duberg2020ufomap} as the map representation is also expected to enhance performance.


\section{ACKNOWLEDGEMENT}
\label{sec:acknowledgement}
The authors thank the MRS group at Czech Technical University, LARICS lab at the University of Zagreb, Smart Autonomous Robotics group at Southern University of Science and Technology and ISR group at Universidade de Lisboa for making their work open for research. Large Language Models (LLMs) have been employed to refine the phrasing and clarity of parts of the written text.

\printbibliography

\end{document}